\documentclass[11pt]{article}
\usepackage{ipac}
\usepackage{times}
\usepackage{url}
\usepackage{latexsym}
\usepackage[icelandic]{babel}
\usepackage[T1]{fontenc}
\usepackage[multiple]{footmisc}

\aclfinalcopy % Uncomment this line for the final submission

\title{Icelandic Parallel Abstracts Corpus}

\author{
  Haukur Barri Símonarson \\
    Miðeind ehf\\
  {\tt haukur@mideind.is } \\ 
  \And
  Vésteinn Snæbjarnarson \\
  Miðeind ehf\\
  {\tt vesteinn@mideind.is}
  }
\date{}

\begin{document}
\maketitle
\begin{abstract}
We present a new Icelandic\textendash{}English parallel corpus, the Icelandic Parallel Abstracts Corpus (IPAC), composed of abstracts from student theses and dissertations. 
The texts were collected from the \emph{Skemman}\footnote{\tt{https://skemman.is}} repository which keeps records of all theses, dissertations and final projects from students at Icelandic universities. 
The corpus was aligned based on sentence-level BLEU scores, in both translation directions, from NMT models using Bleualign. 
The result is a corpus of 64k sentence pairs from over 6 thousand parallel abstracts.
\end{abstract}

\section{Introduction}
Parallel text corpora are the cornerstone of machine translation systems. While recent developments have reduced this dependence somewhat with 
unsupervised neural machine translation they continue to play an important role, 
in particular during fine-tuning \cite{artetxeEffectiveApproachUnsupervised2019a, lampleCrosslingualLanguageModel2019c}.

Parallel data is also of high importance for automatic evaluation of machine translation models via computable metrics such as BLEU or NIST.
Parallel corpora can also be used to automatically construct parallel glossaries or dictionaries.

Manual creation of parallel corpora is time consuming and expensive and naturally occurring texts are thus of great interest. 
For texts in Icelandic and English, one such source is the \mbox{\emph{Skemman}}~repository, 
which contains a collection of student theses and dissertations from all Icelandic universities, including some research papers from faculty. It has been hosted at the National Library of Iceland since 2008 and lists over 35,000 entries.\footnote{Checked on February 8, 2021.}
In this work we gather all available files from the repository, locate and extract parallel abstracts and align the resulting segments.

% mætti mögulega vera sitt eigið subsection einhvers staðar líka

\subsection{Existing corpora}
% tala meira um galla opensubtitles eða vísa í grein sem fjallar um slíkt (t.d. samanburður TED talks vs OpenSubtitles)
%Current parallel data is either of varying quality for one reason or another such as OpenSubtitles, Tatoeba, the Icelandic sagas and Gutenberg literature or 
%is very domain specific (EEA, EMEA, software localizations) or uses terminology or vocabulary that may not be desirable such as religious texts (JW300, the Bible).
%It has therefore been difficult to automatically evaluate the broader generalization performance of existing translation models.

Most of the currently aligned parallel Icelandic\textendash{}English data is found in the ParIce collection of parallel corpora \cite{barkarsonCompilingFilteringParIce2019}. While extensive, the sub-corpus quality is either varying in quality or very domain-specific. Crowd-sourced datasets include OpenSubtitles and Tatoeba while the Icelandic sagas and Gutenberg literature often contain arcane language. The higher quality parallel data is mainly sourced from translated EEA-regulations, medicinal information (EMEA) or software localizations (Ubuntu and KDE). Other datasets contain vocabulary that may not be desirable such as religious texts (Jehova's Witnesses corpus, JW300 \cite{agicJW300WideCoverageParallel2019} and the Bible).
It has therefore been difficult to automatically evaluate the broader generalization performance of existing translation models, something we hope to address with the wide scope and high quality of IPAC.

\begin{table}[h]
\begin{center}
\begin{tabular}{|l|r|}
\hline \bf Corpus & \bf Size \\ 
\hline
The Bible &  33k \\
EEA regulatory texts & 1,700k \\
EMA & 404k \\
European Space Observatory (ESO) & 12.6k \\
OpenSubtitles &  1,300k \\
Tatoeba & 10k \\
Jehova's Witnesses (JW300) & 527k \\
Other$^\ast$ & 93k \\
\hline
\textbf{IPAC} (this work) & 64k \\
\hline
\end{tabular}
\end{center}
\begin{center}
Table 1: Existing parallel corpora, \emph{other$^\ast$} denotes software localizations, Project Gutenberg literature and the Icelandic sagas. \
\end{center}
\end{table}

Multilingual sources with more than 2 languages include Jehova's Witnesses corpus (JW300) and the European Medicines Association corpus (EMA/EMEA). For more information on the aforementioned corpora see \cite{tiedemannParallelDataTools2012} and \cite{barkarsonCompilingFilteringParIce2019}.

\section{Abstract extraction}
A small scraper was written in Python to download the PDF files belonging to each thesis entry. 
At the time of gathering a total of 31k PDF files were collected.
This was necessary because not all entries included abstracts in their metadata, 
and only a small subset included abstracts in more than one language.

The repository provides a functionality for locking documents, optionally with a release date (potentially years in the future). Fortunately, most documents are not locked. Even so, many authors do not use this functionality at all and opt to encrypt their PDF with a password. While it seems the universities encourage or require abstracts in both English and Icelandic, not all documents include both regardless of the language of the document itself.

\subsection{Language detection}
The universities accepts theses, dissertations and final projects in many languages, not just English and Icelandic.
Unfortunately, the language of the main document is not part of the provided metadata, which only denotes the language of the provided abstract or title.
A language might be listed in the keywords, but that was not a reliable indicator of the document language 
(especially so for Icelandic and English, which are usually implicit). Language detection based on abstract related keywords was used.

%%%%%%%%% Not needed?
%\subsection{Document targeting}
%Most entries in the repository list more than one file, only one of which is the main text of the document and the rest might be the front page as a separate file,
%appendices or bibliographies or other project specific files. To limit false positives we attempt to identify which document to extract text from using document %descriptions and some simple heuristics. 

%Using document size is not reliable since many appendices or project specific files are larger than the main document (e.g. they include more images).
%Each file has a description field in the metadata, but it was only filled out some of the time.
%Many authors tend to use standard terms to denote the document content and a mix of simple rules, including 
%but not limited to common terms for the main document filename, 
%excluding files that have an unrelated description in the provided field, and defaulting to the largest document.

\subsection{Text extraction}
The text was extracted from the PDF files with the {\tt pdftotext}\footnote{Part of Poppler,\\ \tt{https://poppler.freedesktop.org/}} software and various ad-hoc rules were written to determine the locations of the abstracts, such as via section title synonyms, length limits, lines starting on lowercase characters, comparable total lengths as well as the assumption that abstracts should occur near the beginning of a given file. A total of 7845 parallel abstracts were found.
   
\section{Alignment}
Sentence segmentation for Icelandic was performed with the Miðeind {\tt Tokenizer} \cite{thorsteinssonWideCoverageContextFreeGrammar2019} for Icelandic as well as English, as it was found to be accurate enough for both.
NMT models were trained over a dataset composed of the pre-existing parallel corpora and backtranslated monolingual text from the news section of the Icelandic Gigaword Corpus (IGC) \cite{steingrimssonRisamalheildVeryLarge2018} and the English section of the newscrawl 
corpus\footnote{{\tt{}http://data.statmt.org/news-crawl}}\footnote{\textasciitilde{}15 million lines were sourced from each language from the backtranslations provided at {\tt https://repository.clarin.is/repository/xmlui/\\handle/20.500.12537/70}.}.
The models were then used to translate their respective source language side of the abstracts. 
{\tt Bleualign}\footnote{{\tt{}https://github.com/rsennrich/Bleualign}}, an implementation of the algorithm described in \cite{sennrichIterativeMTbasedSentence2011}, 
was then used to align the texts, leveraging the output of the NMT models in both translation directions. 

\begin{table}[h]
\begin{center}
\begin{tabular}{|lrr|r|}
\hline \bf Lang. & \bf Segm. & \bf Tokens pre & \bf Tokens after \\ 
\hline
Icelandic &  84694 & 1656k & 1324k\\
English & 83281 & 1811k & 1483k \\
\hline
Aligned & 63870 & \textemdash & \textemdash \\
\hline
\end{tabular}
\end{center}
\begin{center}
Table 2: Alignment results
\end{center}
\end{table}

For a given document one or both translation directions can be used to compute alignments, however when both directions are provided the
intersection of the alignments from each direction is used instead. The end result is a higher precision alignment, at the cost of lower recall.
% using only one translation direction (eng->isl), the aligned sentences were 75066

\begin{table}[h]
\begin{center}
\begin{tabular}{|l|rrr|}
\hline \bf Field & \bf Abstracts & \bf Acc. & \bf Rej.  \\ 
\hline
Social sciences & 2369 & 23962 & 27502  \\
Natural sciences & 1248 & 11886 & 13884 \\
Medical and health & 1195 & 15546 & 17388 \\
Humanities & 1026 & 10045 & 11455 \\
Business & 604 & 5355 & 6095 \\
Misc & 193 & 1723 & 1982\\
\hline
\end{tabular}
\end{center}
\begin{center}
Table 3: Domain origin of parallel sentences
\end{center}
\end{table}

The grouping in Table 3 is based on the school or department where a given thesis was submitted. Note that the input here are individual sentences as opposed to the segments in Table 2 where some sentences may have been joined into a single many-to-one alignment.

\subsection{Alignment quality}
Due to the abstracts being written without any constraints of sentence-to-sentence translation some of them were found to align poorly due to content
being omitted in either language. Fortunately most abstracts were almost translated at the sentence level and align well.

A random sample of 100 pairs was selected for manual evaluation and were classified into 4 categories: a) correct , b) near correct (slight loss of meaning, different choice of words), c) partial (some meaning completely lost, i.e. part of sentence gone or added). None were completely wrong, i.e. no alignment was present, as shown in Table 4.

\begin{table}[h]
\begin{center}
\begin{tabular}{|l|r|}
\hline \bf Group & \bf \%  \\ 
\hline
Correct & 71 \% \\
Near correct & 22 \% \\
Partial & 7 \% \\
Incorrect & 0 \% \\
\hline
\end{tabular}
\end{center}
\begin{center}
Table 4: Human evaluation
\end{center}
\end{table}

%\section{Discussion}
\section{Discussion and future work}
Most of the heuristics applied in the extraction and filtering stage were unnecessarily coarse for the sake of eliminating noise and increasing precision. 
More fine-grained heuristics and newer translation models may significantly increase the total yield without introducing much additional noise in future versions.
Uncertainty estimation \cite{fomichevaUnsupervisedQualityEstimation2020} may also be able to identify poor alignments more accurately than any heuristics-based approach.
BLEU is also a poor metric at the sentence-level and is typically used at the corpus level, 
a translation metric such as BERTScore \cite{zhangBERTScoreEvaluatingText2020} which is a closer correlate with human judgement may help in this regard.

\subsection{Release}
The shuffled aligned parallel corpus is made availble on the CLARIN-repository\footnote{\tt{https://repository.clarin.is}} in pre-defined splits.\footnote{Pending submission review.}

\section{Conclusion}
We have extracted and aligned a high quality parallel Icelandic\textendash{}English corpus IPAC from a wide variety of academic fields.
It is orthogonal to other Icelandic\textendash{}English parallel corpora and consists of a wide variety of topics.
We envision it serves well, not only for training, but also as a much welcome benchmark of Icelandic\textendash{}English machine translation systems and look forward to seeing it in use.

\bibliographystyle{acl_natbib}
\bibliography{nodalida2021}

\end{document}